\documentclass{bvm} 
\usepackage[nolist]{acronym}
\usepackage{multirow}
\usepackage{graphicx}

\begin{document}

\newcommand{\bvmyear}{2026}
\selectlanguage{english} 


\title{SWAN - Enabling Fast and Mobile Histopathology Image Annotation through Swipeable Interfaces}



\titlerunning{SWAN - SWipeable ANnotations}

\author{
	\fname{Sweta} \lname[0000-0001-5029-5378]{Banerjee} \inst{1} \affiliation{Flensburg University of Applied Sciences, Flensburg, Germany} \authorsEmail{sweta.banerjee@hs-flensburg.de} \isResponsibleAuthor,
	\fname{Timo} \lname{Gosch} \inst{1} \equalContribution \affiliation{Flensburg University of Applied Sciences, Flensburg, Germany} \authorsEmail{timo.gosch@stud.hs-flensburg.de},
    \fname{Sara} \lname{Hester} \inst{1} \equalContribution \affiliation{Flensburg University of Applied Sciences, Flensburg, Germany} \authorsEmail{sara.hester@stud.hs-flensburg.de},
    \fname{Viktoria} \lname[0009-0002-0062-844X]{Weiss} \inst{2} \affiliation{University of Veterinary Medicine, Vienna, Austria} \authorsEmail{Viktoria.Weiss@vetmeduni.ac.at},
    \fname{Thomas} \lname{Conrad} \inst{3} \affiliation{Freie Universit\"{a}t Berlin, Berlin, Germany} \authorsEmail{thomas.conrad@fu-berlin.de},
    \fname{Taryn A.} \lname[0000-0001-5740-9550]{Donovan} \inst{4} \affiliation{Schwarzman Animal Medical Center, New York, USA} \authorsEmail{Taryn.Donovan@amcny.org},
    \fname{Nils} \lname[0009-0005-3672-9316]{Porsche} \inst{1} \affiliation{Flensburg University of Applied Sciences} \authorsEmail{nils.porsche@hs-flensburg.de},
    \fname{Jonas} \lname[0000-0002-0335-1194]{Ammeling} \inst{5} \affiliation{Technische Hochschule Ingolstadt, Ingolstadt, Germany} \authorsEmail{Jonas.Ammeling@thi.de},
    \fname{Christoph} \lname{Stroblberger} \inst{6} \affiliation{Medical University of Vienna, Vienna, Austria} \authorsEmail{christoph.stroblberger@meduniwien.ac.at},
    \fname{Robert} \lname{Klopfleisch} \inst{3} \affiliation{Freie Universit\"{a}t Berlin, Berlin, Germany} \authorsEmail{robert.klopfleisch@fu-berlin.de},
	\fname{Christopher} \lname{Kaltenecker} \inst{6} \affiliation{Medical University of Vienna, Vienna, Austria} \authorsEmail{christopher.kaltenecker@meduniwien.ac.at},
	\fname{Christof A.} \lname[0000-0002-2402-9997]{Bertram} \inst{2} \affiliation{University of Veterinary Medicine, Vienna, Austria} \authorsEmail{Christof.bertram@vetmeduni.ac.at},
	\fname{Katharina} \lname[0000-0001-7600-5869]{Breininger} \inst{7} \affiliation{Julius-Maximilians-Universit\"{a}t W\"{u}rzburg, W\"{u}rzburg, Germany} \authorsEmail{katharina.breininger@uni-wuerzburg.de},
	\fname{Marc} \lname[0000-0002-5294-5247]{Aubreville} \inst{1} \affiliation{Flensburg University of Applied Sciences, Flensburg, Germany} \authorsEmail{marc.aubreville@hs-flensburg.de}
}

\authorrunning{Banerjee et al.}

\institute{
\inst{1} Flensburg University of Applied Sciences, Flensburg, Germany\\
\inst{2} University of Veterinary Medicine, Vienna, Austria\\
\inst{3} Freie Universit\"{a}t Berlin, Berlin, Germany\\
\inst{4} Schwarzman Animal Medical Center, New York, USA\\
\inst{5} Technische Hochschule Ingolstadt, Ingolstadt, Germany\\
\inst{6} Medical University of Vienna, Vienna, Austria\\
\inst{7} Julius-Maximilians-Universit\"{a}t W\"{u}rzburg, W\"{u}rzburg, Germany
}

\email{sweta.banerjee@hs-flensburg.de}

\maketitle

\begin{abstract}
	The annotation of large-scale histopathology image datasets remains a major bottleneck in developing robust deep learning models for clinically relevant tasks, such as mitotic figure classification. Folder-based annotation workflows are usually slow, fatiguing, and difficult to scale. To address these challenges, we introduce SWipeable ANnotations (SWAN), an open-source, MIT-licensed web application that enables intuitive image patch classification using a swiping gesture. SWAN supports both desktop and mobile platforms, offers real-time metadata capture, and allows flexible mapping of swipe gestures to class labels. In a pilot study with four pathologists annotating 600 mitotic figure image patches, we compared SWAN against a traditional folder-sorting workflow. SWAN enabled rapid annotations with pairwise percent agreement ranging from 86.52 \% to 93.68 \% (Cohen’s $\kappa$ = 0.61–0.80), while for the folder-based method, the pairwise percent agreement ranged from 86.98 \% to 91.32 \% (Cohen’s $\kappa$ = 0.63–0.75) for the task of classifying atypical versus normal mitotic figures, demonstrating high consistency between annotators and comparable performance. Participants rated the tool as highly usable and appreciated the ability to annotate on mobile devices. These results suggest that SWAN can accelerate image annotation while maintaining annotation quality, offering a scalable and user-friendly alternative to conventional workflows. The full codebase of the SWAN project, including analyses, can be found here: \url{https://anonymous.4open.science/r/SWAN-BD51}.
\end{abstract}

\section{Introduction}

The digitization of \ac{WSIs} has enabled the development of deep learning models for automating clinically relevant tasks. 
However, these models are dependent on accurate annotations by domain experts. In medical imaging, and particularly in histopathology, generating high-quality labels is often a major bottleneck, as it requires domain experts to inspect thousands of images, a process that is time-intensive, tiring, and subject to inter-observer variability. Similar challenges also exist in other fields that rely on patch-level image classification, such as cell biology, materials science, etc., where large annotated datasets are essential but difficult to produce at scale. 
In digital histopathology research, a common annotation workflow involves manually sorting image patches into different folders. While simple, this approach is not optimized for speed, scalability, completeness, or user engagement. Although a wide range of annotation tools have been developed in other domains~\cite{maree2016collaborative, bankhead2017qupath}, many of which are commercial, most are tailored for desktop computer environments, limiting flexibility and accessibility. Recent advances in mobile computing and cloud-based data management offer new opportunities to rethink how expert annotation is performed. Mobile-enabled interfaces can allow domain experts to contribute annotations conveniently, even outside the traditional laboratory environment, streamlining collaborative labeling.

In this work, we present \ac{SWAN}, a MIT-licensed, mobile-friendly annotation platform designed to streamline patch-level labeling in digital histopathology. The system supports rapid, scalable, and user-friendly workflows, enabling efficient annotation directly from smartphones or tablets. We describe the platform’s design principles, implementation, and \ac{UI}, and evaluate its usability, annotation consistency and speed compared to conventional desktop-based methods. Our results demonstrate that mobile-enabled annotation can significantly enhance flexibility and engagement in expert-driven image labeling, addressing a key barrier to scalable dataset generation in computational pathology.

\begin{figure}[b]
  \centering
  \includegraphics[width=0.18\textwidth,trim={0 20 0 100},clip]{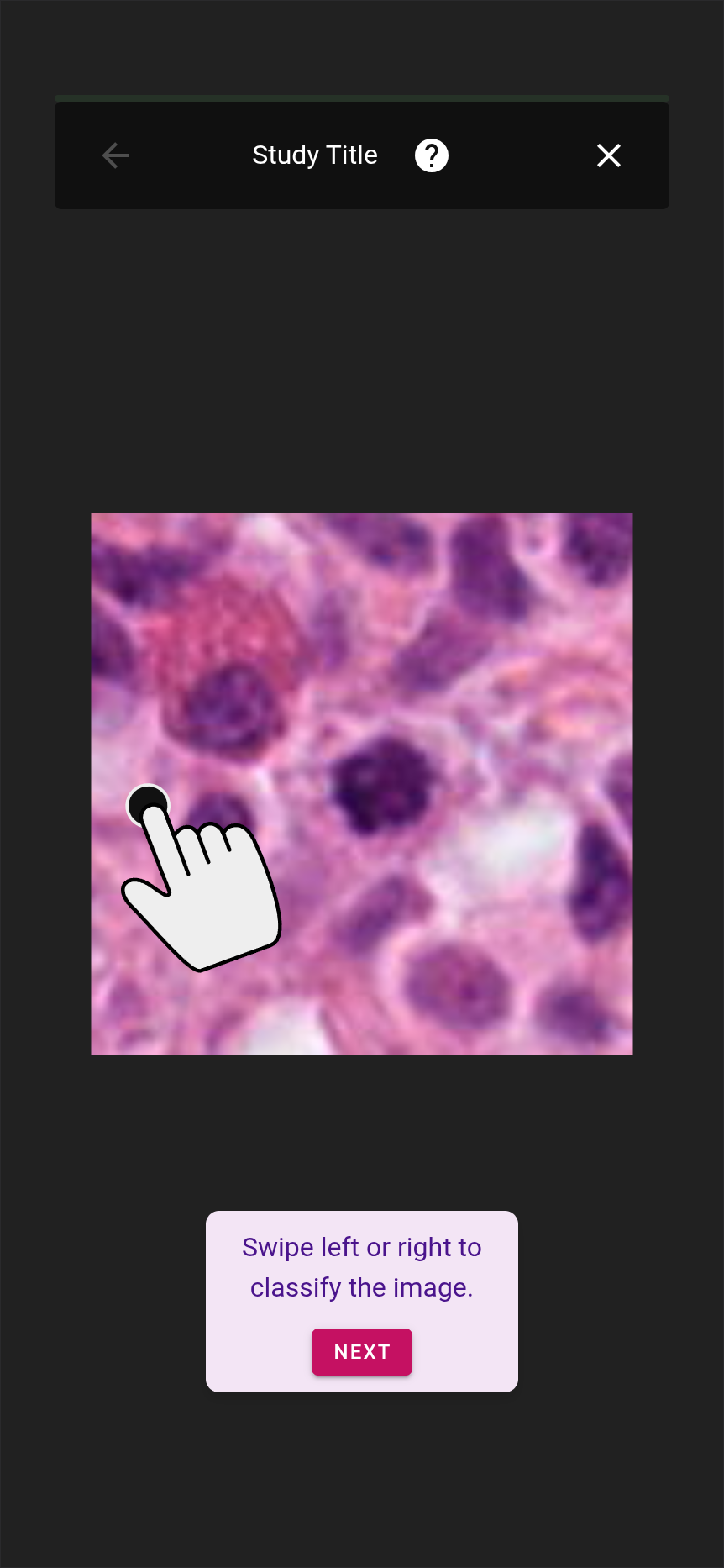}\hfill
  \includegraphics[width=0.18\textwidth,trim={0 20 0 100},clip]{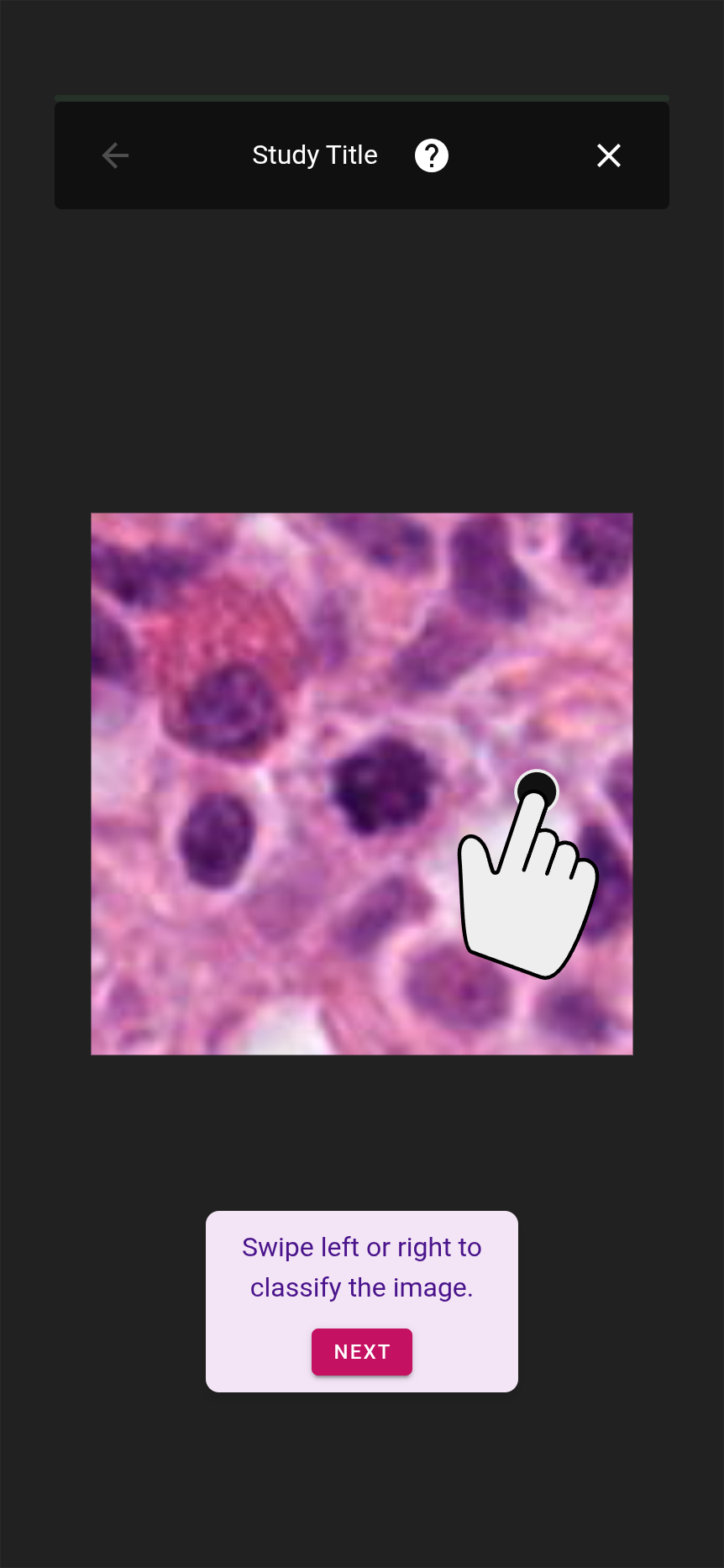}\hfill
  \includegraphics[width=0.18\textwidth,trim={0 20 0 100},clip]{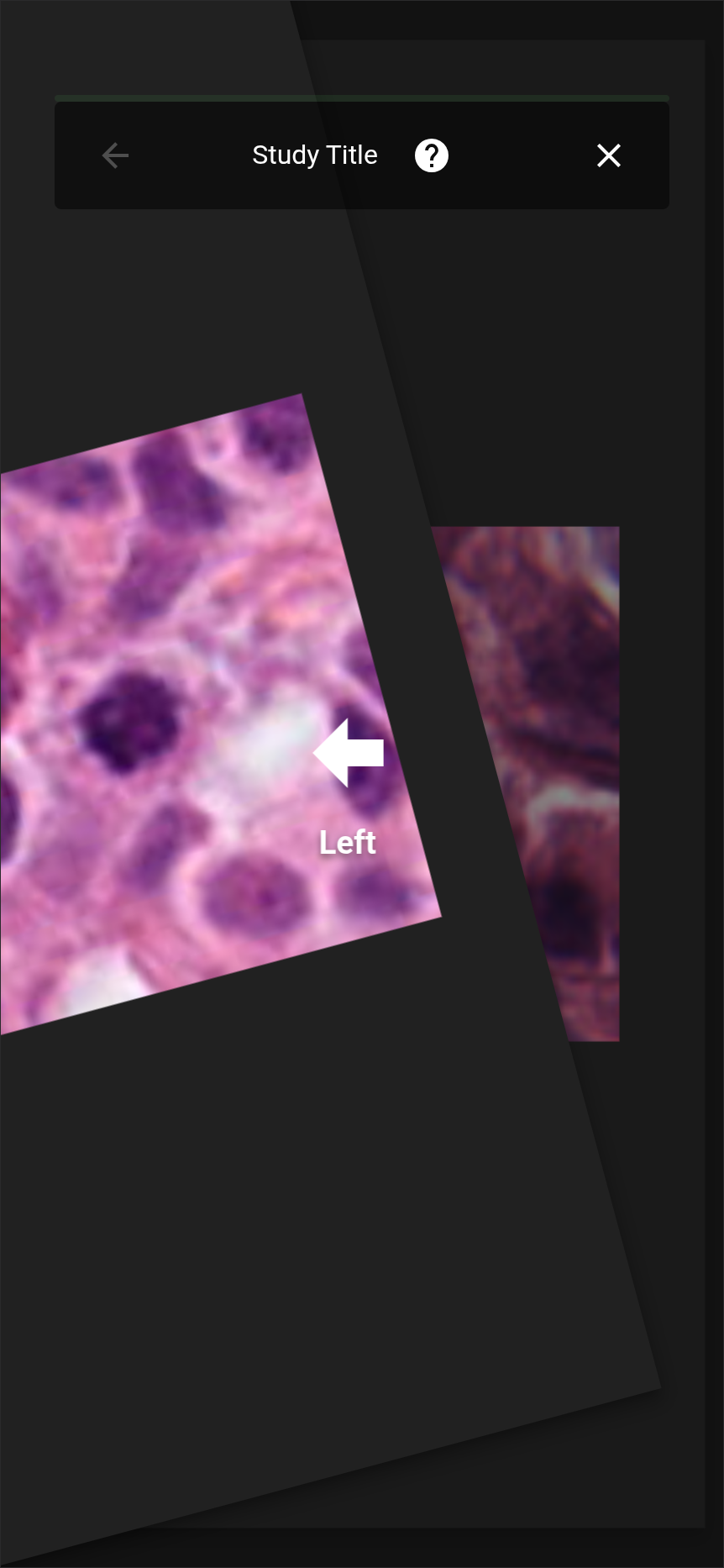}\hfill
  \includegraphics[width=0.18\textwidth,trim={0 20 0 100},clip]{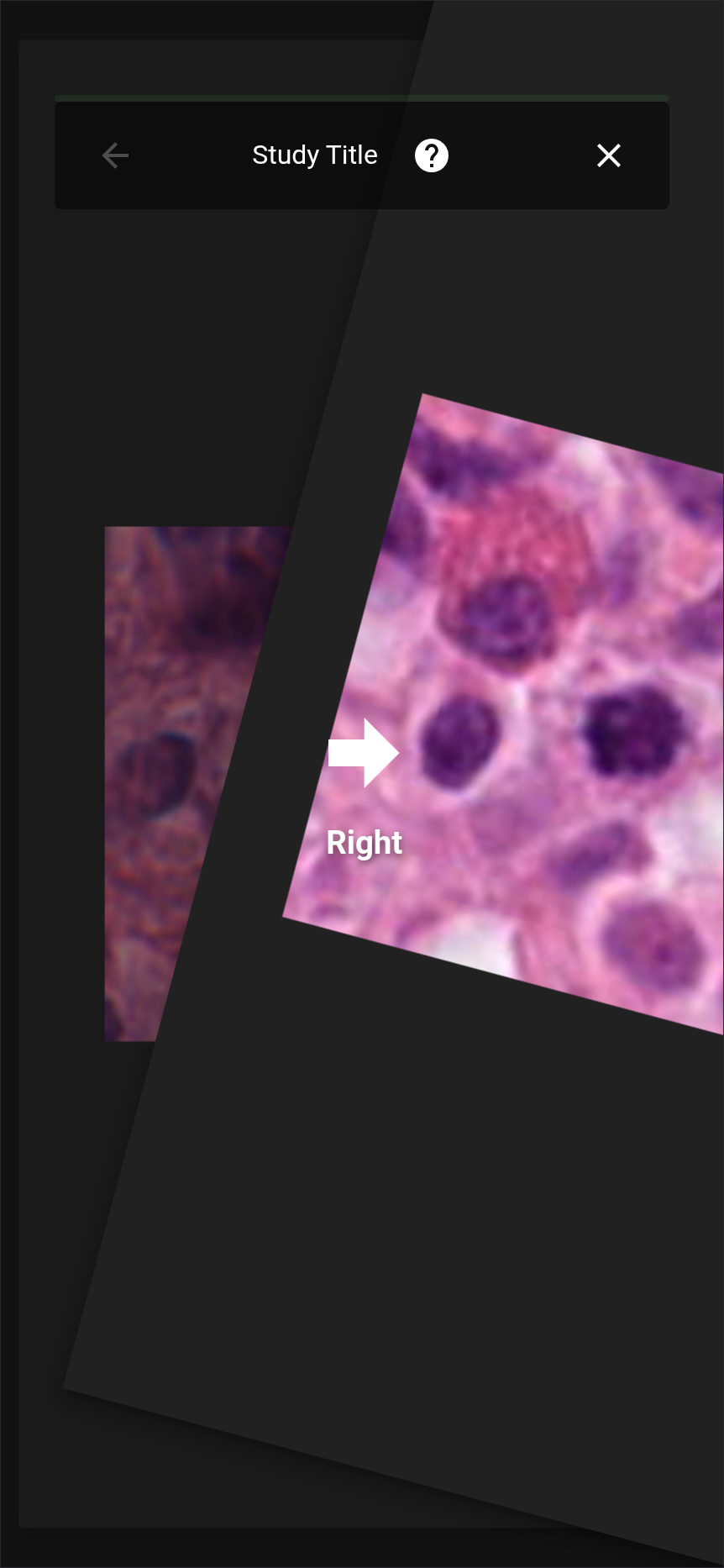}\hfill
  \includegraphics[width=0.18\textwidth,trim={0 20 0 100},clip]{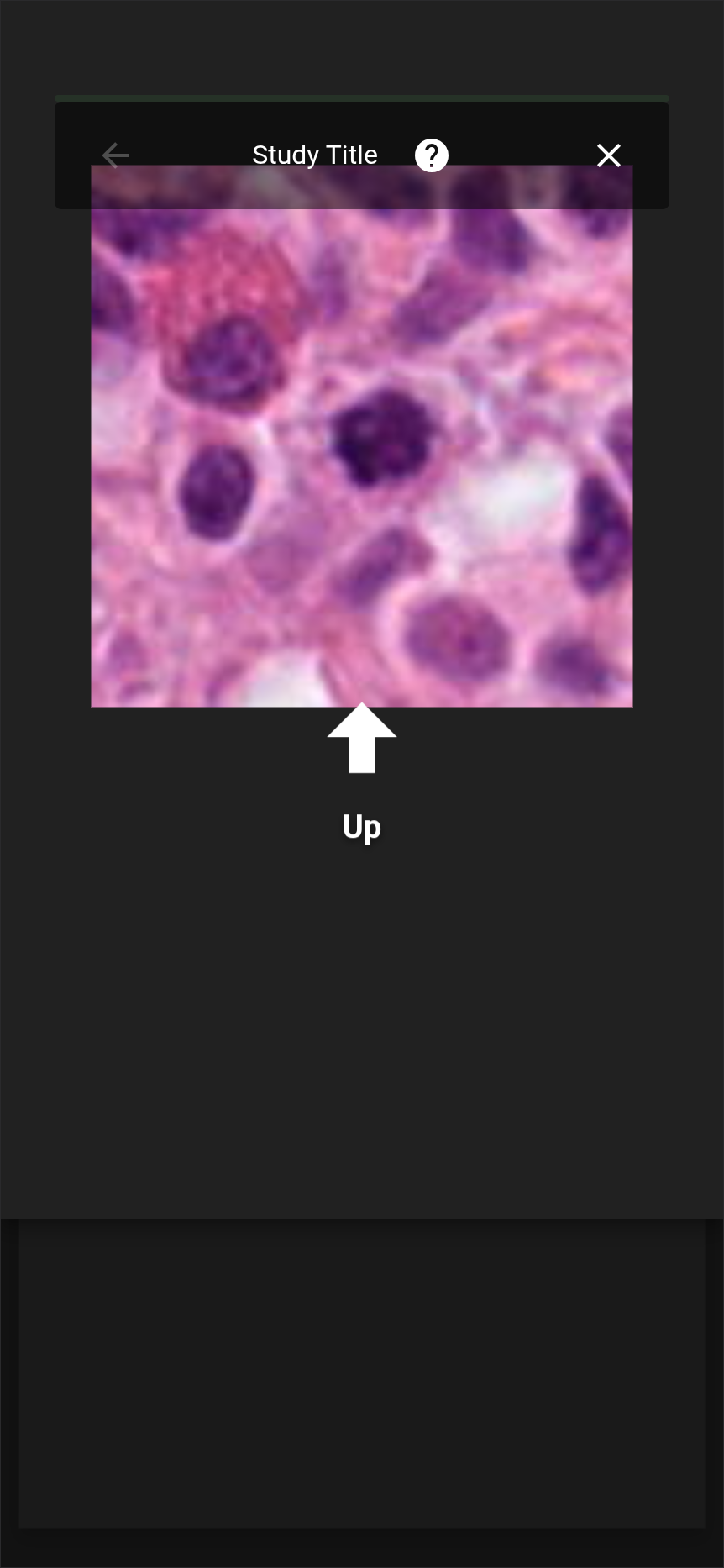}\hfill
  \caption{Swipe interactions in SWAN (mobile version): left, right, and up for classification or postponing, and corresponding thumb icons from the help intro.}
  \label{fig:swipe-interface}
\end{figure}



\section{Materials and Methods}

\begin{SCfigure}[10][b]
    \centering
    \includegraphics[width=0.50\figwidth]{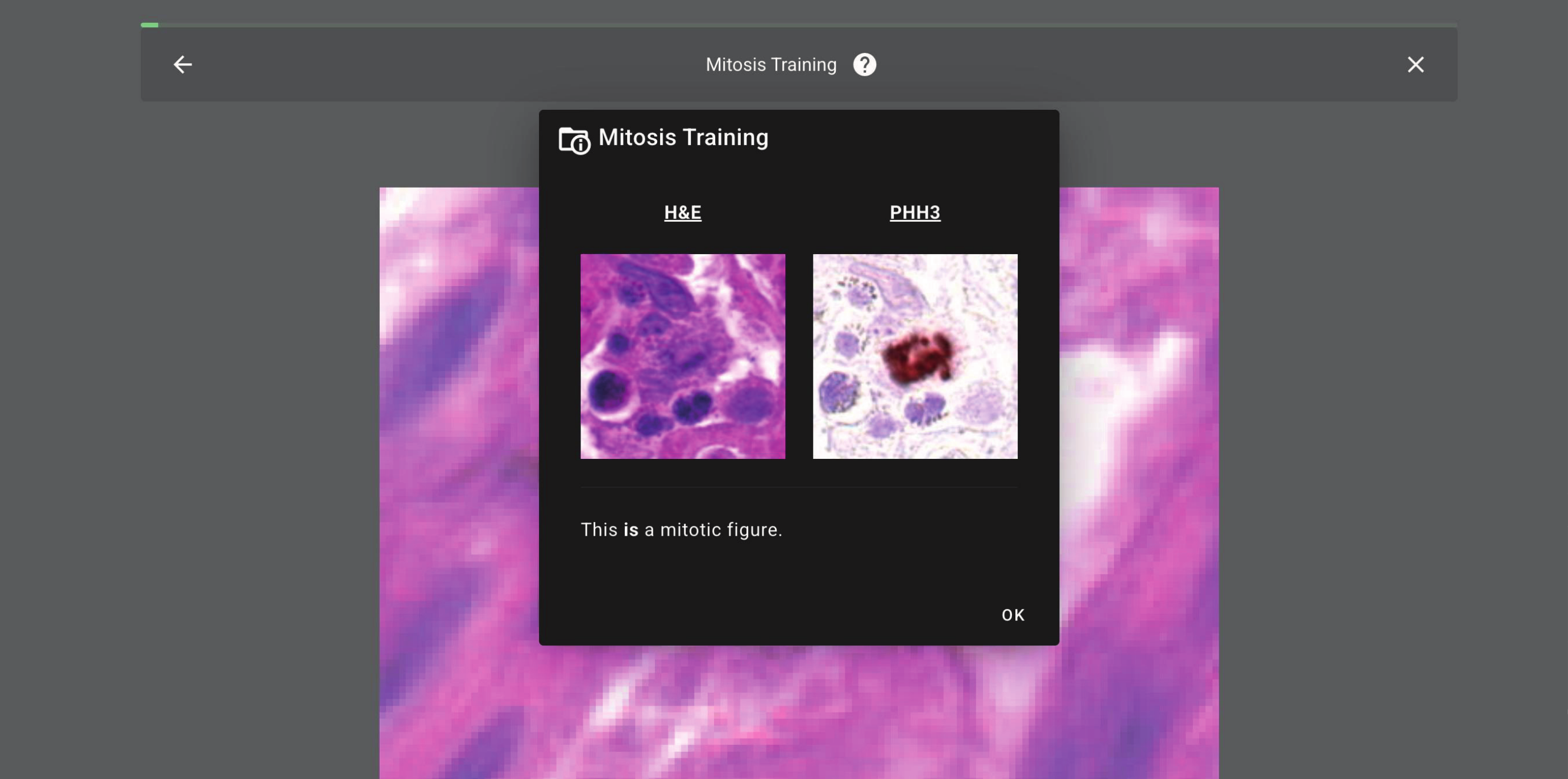}
    \caption{Application of SWAN for pathologist training (desktop version).}
    \label{fig:path-edu}
    \altText{Diagram showing the use of the SWAN annotation tool by pathologists during training (desktop version).}
\end{SCfigure}

Using a swipe-based interaction, popularized for the use-case of binary decision making by dating apps, our tool makes the annotation process more efficient and engaging. Users can label image patches with ease by simply swiping left, right, up, or down. Mistakes can be corrected by going back. The interface can also be used on a tablet or on a desktop computer with a mouse. The tool records timestamped user input and metadata, which enables smooth integration with current annotation pipelines and quantitative analyses. We envision two major use cases for this tool:
\begin{itemize}
    \item \textbf{Annotation}: For annotation of images patches into up to four categories, by swiping left, right, up and down (Fig. \ref{fig:swipe-interface}).
    \item \textbf{Pathologist education}: Training pathologists by displaying the correct option after they swipe their choice (Fig. \ref{fig:path-edu}).
\end{itemize}

\subsection{Software Architecture and Workflow Design}
\label{0000-sec-payment}

\ac{SWAN}, which is available on our github repository\footnote{\url{https://anonymous.4open.science/r/SWAN-BD51}}, is implemented using a Django-based backend and a Vuetify-based frontend. 
It has two main \acp{UI}:
\begin{itemize}
    \item \textbf{Administrator:} Enables study definition, dataset management, and user-management. Annotation classes for each direction and \ac{UI}-related settings can be configured here. 
    \item \textbf{Participant:} Provides access to assigned studies for annotation and training.
\end{itemize}

\noindent
A typical annotation workflow then involves the following steps:
\begin{enumerate}
    \item The admin creates a project group and assigns the existing experts to it.
    \item The admin uploads the data set as an image archive (\texttt{.zip} or \texttt{.tar} format).
    \item The admin creates the study, selects the dataset and defines the \ac{UI} configuration.
    \item Participants log in and open an assigned study from the overview. New users are greeted with an onboarding sequence.
    \item Participants annotate by swiping left, right, up, or down, and can resume from their last position in subsequent sessions.
    \item The annotations and associated metadata can be exported to a CSV file.
\end{enumerate}
The order of the data set is randomized for each participant.

\subsection{User Study}
To demonstrate the practical use of \ac{SWAN}, we conducted a pilot user study, evaluating the usability and efficiency of \ac{SWAN}’s swipe-based interface for a classification annotation workflow and to compare it against the routine desktop folder-based annotation workflow. 
We chose the task of subclassification of \acp{MF} into typical and atypical \acp{MF}. This task is particularly well-suited for evaluating the impact of different annotation strategies because it is an emerging topic of research and has shown only moderate inter-rater variability in prior studies \cite{bertram2025histologic}. We chose a random subset of 600 mitotic figure image patches that were cropped from the canine cutaneous mast cell tumor (CCMCT) dataset \cite{bertram2019large}. All images were in the~\texttt{.png} format with dimensions of $128 \times 128$ pixels. 


Four pathologists participated in this user study. To minimize recall bias, a washout period of at least two weeks was implemented before each phase. The user study was conducted in three phases:

\paragraph{\ac{SWAN} (Initial)}
In the first of three phases, participants were provided with login credentials to access the \ac{SWAN} platform and annotate the assigned image set. For this study, the left swipe was mapped to the \textit{normal} class and the right swipe to the \textit{atypical} class. 
These annotations, including timestamp information, were then exported as CSV files for downstream quantitative analyses. 

After completing the annotations, each participant was asked to fill in a post-study questionnaire that collected information about their professional background, experience in histopathology, familiarity with \acp{AMF} classification, and device used. Usability aspects like responsiveness, ease of use, intuitiveness of the swipe mechanism, perceived image quality, confidence, engagement, and willingness to reuse, were assessed using 5-point Likert scales. Additional open-ended questions addressed uncertainty during classification, technical issues, and suggestions for improvement.

\paragraph{\ac{SWAN} (Enhanced)}

In the second study phase, we incorporated feedback from pathologists and updated the \ac{SWAN} app as follows:
\begin{itemize}
    \item To address the pixelation reported in the first phase, we introduced two features: an image-scaling option to display images at partial screen size and a toggle for image interpolation.
    \item We introduced a feature allowing to assign a swiping direction, e.g., upwards, with a postpone functionality, enabling revisiting difficult or confusing images at the end of the trial. 
\end{itemize}

Upon finishing the annotations, each participant was asked to complete the post-study questionnaire again, where they were additionally asked focused questions regarding factors related to new functionality in this phase. 

\paragraph{Folder-Based Method}
In this phase, the same participants were asked to annotate the same image set using their routine desktop folder-based method, where they put the images into two different folders according to normal or atypical \acp{MF} and to manually record their annotation times. 

\section{Results}
\label{0000-sec-latex-template}


For the initial \ac{SWAN} phase, mean annotation times per image ranged from 1.58\,s to 2.83\,s across participants, while for the enhanced \ac{SWAN} phase, mean annotation times ranged from 1.67\,s to 2.60\,s. For the folder-based method, the participants were generally slower, with mean annotation times ranging from 1.83\,s to 4.30\,s (Tab. \ref{tab:participant-summary}). Pairwise percent agreement for the initial phase of \ac{SWAN} ranged from 81.47\,\% to 87.98\,\%, with Cohen's $\kappa$ values between 0.55 and 0.67. For the enhanced version of \ac{SWAN}, the pairwise percent agreement during the \ac{SWAN} phase ranged from 86.52 \% to 93.68 \%, with Cohen’s $\kappa$ values between 0.61 and 0.80 (Tab. \ref{tab:pairwise-kappa}). For the folder-based method, agreement was comparable to the enhanced \ac{SWAN}, with percent agreement ranging from 86.98 \% to 91.49 \% and $\kappa$ values between 0.63 and 0.75. Fleiss’ $\kappa$ across all four participants was 0.610 for the initial \ac{SWAN} phase, 0.680 for the enhanced \ac{SWAN} phase and 0.695 for the folder-based phase. 

\begin{table}[t]
\centering
\small
\setlength{\tabcolsep}{12pt}
\renewcommand{\arraystretch}{1.2}
\caption{Average annotation times in seconds per participant per image across all three phases.}
\label{tab:participant-summary}
\begin{tabular}{lccc}
\hline
\textbf{Participant} & \textbf{SWAN(Initial)} & \textbf{SWAN(Enhanced)} & \textbf{Folder-Based} \\
\hline
expert 1 & 1.58 & 1.67 & 1.83 \\
expert 2 & 2.83 & 2.41 & 4.30 \\
expert 3 & 2.72 & 2.60 & 2.25 \\
expert 4 & 1.67 & 2.18 & 2.90 \\
\hline
\end{tabular}
\end{table}


\begin{table}[t]
\centering
\small
\caption{Pairwise inter-observer agreement across participants for SWAN~(Initial), SWAN~(Enhanced), and folder-based annotation modes. Percent agreement and Cohen's $\kappa$ are reported separately for each mode.}
\label{tab:pairwise-kappa}
\resizebox{\textwidth}{!}{%
\begin{tabular}{lcccccc}
\hline
\multirow{2}{*}{\textbf{Participant Pair}} 
& \multicolumn{2}{c}{\textbf{SWAN(Initial)}} 
& \multicolumn{2}{c}{\textbf{SWAN(Enhanced)}} 
& \multicolumn{2}{c}{\textbf{Folder-Based Method}} \\
\cline{2-7}
& \textbf{Agreement} & \textbf{Cohen's $\kappa$}
& \textbf{Agreement} & \textbf{Cohen's $\kappa$}
& \textbf{Agreement} & \textbf{Cohen's $\kappa$} \\
\hline
expert 1 vs expert 2     & 87.98 \% & 0.66 & 86.52 \% & 0.61 & 90.15 \% & 0.69 \\
expert 1 vs expert 3    & 85.64 \% & 0.61 & 87.67 \% & 0.66 & 87.15 \% & 0.63 \\
expert 1 vs expert 4  & 87.81 \% & 0.67 & 93.68 \% & 0.80 & 91.49 \% & 0.73 \\
expert 2 vs expert 3       & 82.30 \% & 0.55 & 87.83 \% & 0.69 & 86.98 \% & 0.66 \\
expert 2 vs expert 4     & 85.14 \% & 0.63 & 87.52 \% & 0.66 & 91.32 \% & 0.75 \\
expert 3 vs expert 4    & 81.47 \% & 0.55 & 87.67 \% & 0.67 & 89.65 \% & 0.72 \\
\hline
\end{tabular}}
\end{table}


In addition to the quantitative analyses, the participants gave us feedback through a post-study questionnaire after each iteration of the \ac{SWAN} study. The questionnaires can be found at our GitHub repository. Overall usability ratings were high across both SWAN phases (\ac{SWAN} (Initial): Mean = 5.00; \ac{SWAN} (Enhanced): Mean = 5.00). High ratings similarly were given for the responsiveness of the smartphone-and-swipe actions (\ac{SWAN}  (Initial) Mean = 4.75; \ac{SWAN} (Enhanced) Mean = 5.00). 
Perceived quality of images improved from Mean = 4.00 during \ac{SWAN}  (Initial), to Mean = 4.75 during \ac{SWAN} (Enhanced), as a result of the adjustable image scaling, which was also rated positively (Mean = 4.00). In \ac{SWAN} (Enhanced), additional questions were included in the questionnaire to examine the new image scaling feature and the increased upward swipe to delay unclassified cases. No significant technical issues were reported across all usability ratings. Participants indicated they would be likely to use the platform to conduct future studies involving larger sample sizes and had the intention to recommend the platform to a colleague.
Overall, participants expressed strong willingness to use the tool for larger datasets and to recommend it to colleagues, indicating good acceptance and usability of the platform.

\section{Discussion}
\label{0000-sec-files}

Annotation can be a repetitive and mentally fatiguing task, and finding ways to make it more engaging is an important step toward scaling annotation efforts. The concept behind SWAN is to provide an annotation experience that is more intuitive, lightweight, and accessible, allowing users to label data not only at a desktop PC but also on mobile devices, in more flexible settings.

Our pilot study demonstrates that this idea has strong potential. The app received positive feedback for its usability and convenience, with improved ratings in the second phase of the study, particularly for image quality and responsiveness.
After introducing image scaling and upward swipe functionality in the enhanced version of \ac{SWAN}, the inter-rater agreement (Cohen’s $\kappa$ 0.61–0.80, Fleiss' $\kappa$ 0.680) was comparable to the folder-based method (Cohen’s $\kappa$ 0.63–0.75, leiss' $\kappa$ 0.695), as shown in Tab. \ref{tab:pairwise-kappa}.

While these small differences could well be explained by measurement error, other factors might also be involved: for one, there was a lower familiarization with the \ac{SWAN} interface. Furthermore, since the \ac{SWAN} tool presents one image at a time, unlike folder-based workflows that provide more contextual overview, a limited comparability between samples might be involved that could reduce discrimination. Future versions of \ac{SWAN} can directly address these limitations by incorporating enhanced navigation options or hybrid viewing modes.

Overall, these results point to a promising starting point - an annotation platform that can make annotation for classification tasks more enjoyable and flexible without compromising the quality. Continued development and larger user studies will help determine how best to balance engagement, efficiency, and annotation quality.

\begin{acknowledgement}
Withheld for blind peer review.
\end{acknowledgement}

\printbibliography

\begin{acronym}
\acro{CNN}{Convolutional Neural Network}
\acro{WSIs}{Whole Slide Images}
\acro{WSI}{Whole Slide Image}
\acro{MF}{mitotic figure}
\acro{AMF}{atypical mitotic figure}
\acro{UI}{user interface}
\acro{SWAN}{SWipeable ANnotations}
\end{acronym}

\end{document}